\documentclass[10pt,twocolumn,letterpaper]{article}

\usepackage[pagenumbers]{cvpr} 
\usepackage{pifont}
\usepackage{listings}
\usepackage{lipsum}
\usepackage{threeparttable}
\usepackage[table]{xcolor}
\usepackage{multirow}

\definecolor{cvprblue}{rgb}{0.21,0.49,0.74}
\usepackage[pagebackref,breaklinks,colorlinks,allcolors=cvprblue]{hyperref}
\def\paperID{2814} 
\def\confName{CVPR}
\def\confYear{2026}

\title{Structure-Aware Prototype Guided Trusted Multi-View Classification}

\author{First Author\\
Institution1\\
Institution1 address\\
{\tt\small firstauthor@i1.org}
\and
Second Author\\
Institution2\\
First line of institution2 address\\
{\tt\small secondauthor@i2.org}
}

\author{
Haojian Huang\textsuperscript{1,2} \quad
Jiahao Shi\textsuperscript{3} \quad
Zhe Liu\textsuperscript{4} \quad
Harold Haodong Chen\textsuperscript{1} \\
Han Fang\textsuperscript{2} \quad
Hao Sun\textsuperscript{2}\thanks{Corresponding author.} \quad
Zhongjiang He\textsuperscript{2} \\
\textsuperscript{1}HKUST(GZ) \quad
\textsuperscript{2}Institute of Artificial Intelligence (TeleAI), China Telecom\\
\textsuperscript{3}Harbin Engineering University \quad
\textsuperscript{4}Universiti Sains Malaysia
}

\begin{document}
\maketitle
\begin{abstract}
Trustworthy multi-view classification (TMVC) addresses the challenge of achieving reliable decision-making in complex scenarios where multi-source information is heterogeneous, inconsistent, or even conflicting. Existing TMVC approaches predominantly rely on globally dense neighbor relationships to model intra-view dependencies, leading to high computational costs and an inability to directly ensure consistency across inter-view relationships. Furthermore, these methods typically aggregate evidence from different views through manually assigned weights, lacking guarantees that the learned multi-view neighbor structures are consistent within the class space, thus undermining the trustworthiness of classification outcomes. To overcome these limitations, we propose a novel TMVC framework that introduces prototypes to represent the neighbor structures of each view. By simplifying the learning of intra-view neighbor relations and enabling dynamic alignment of intra- and inter-view structure, our approach facilitates more efficient and consistent discovery of cross-view consensus. Extensive experiments on multiple public multi-view datasets demonstrate that our method achieves competitive downstream performance and robustness compared to prevalent TMVC methods.
\end{abstract}
\section{Introduction}
\label{sec:introduction}
\begin{figure}[ht]
    \centering
    \includegraphics[width=1\linewidth]{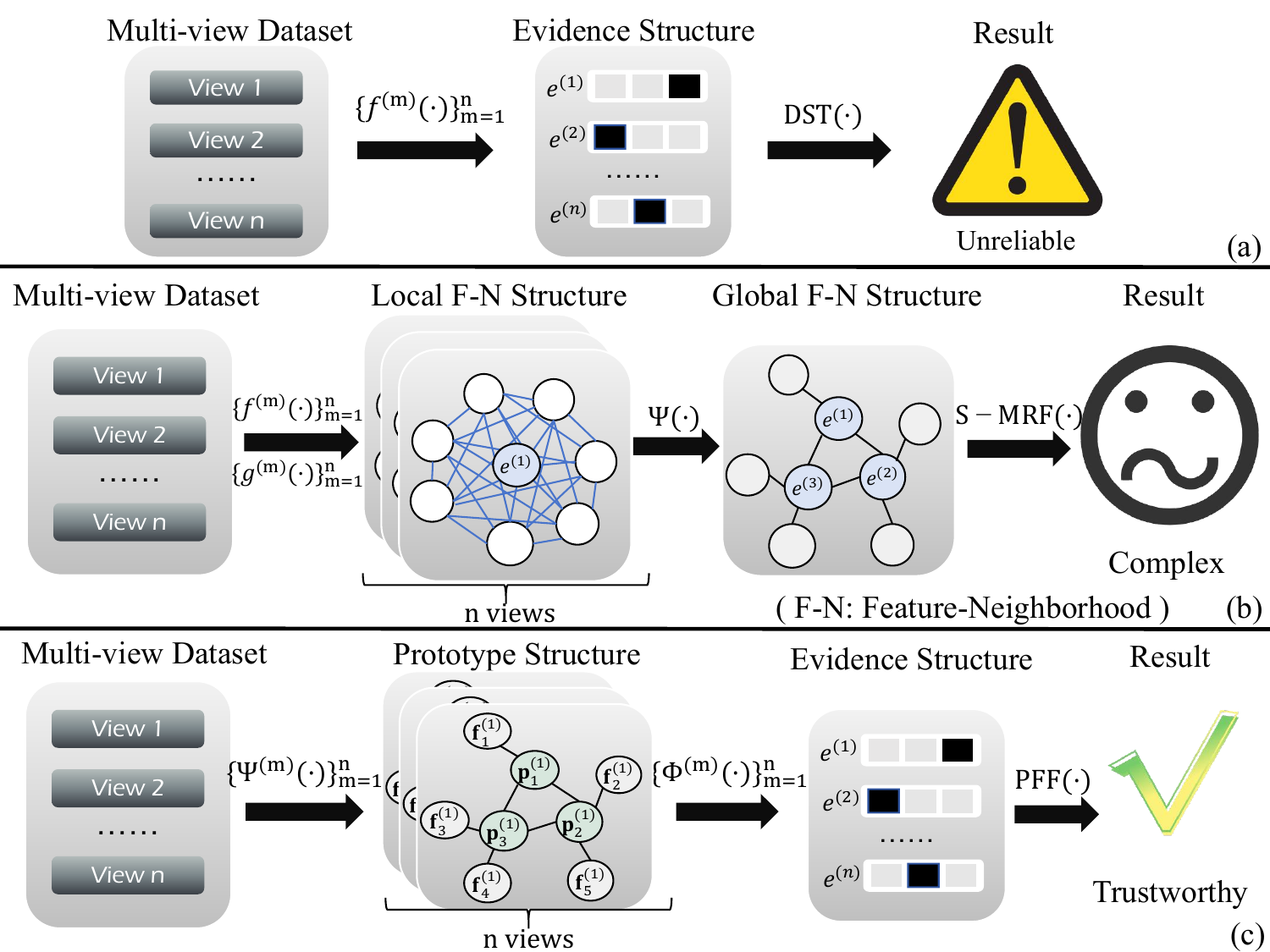}
    \vspace{-2em}
    \caption{Comparison illustration. (a) Traditional TMVC method uses DNNs $\{f^{(m)}(\cdot)\}^{n}_{m=1}$ for evidence extraction but ignores latent neighborhood structures, risking unreliability with conflicting views. (b) TUEND adds feature-neighborhood info via DNNs $\{f^{(m)}(\cdot)\}^{n}_{m=1}$ and GNNs $\{g^{(m)}(\cdot)\}^{n}_{m=1}$, but explicit graph construction raises complexity and limits scalability. (c) Our method focuses on local prototype structures, integrating latent neighborhood structures flexibly in a computation-friendly manner.}
    \label{fig1}
    \vspace{-1.4em}
\end{figure}
Trusted multi-view classification (TMVC) leverages Evidential Deep Learning (EDL)~\citep{sensoy2018evidential}, which is rooted in subjective logic~\citep{josang2016subjective} and Dempster-Shafer theory (DST)~\citep{dempster1968generalization}, to evaluate the quality and uncertainty of each view. This enables more discriminative and reliable decision-making through adaptive multi-view information integration, as demonstrated effective in various fields, such as multimodal analysis~\citep{chen2024finecliper,MCCA}, video understanding~\citep{ma2024beyond,xu2021videoclip,yue2024evidential}, and cross-domain recognition~\citep{huang2024crest,huang2024evidential,bao2021evidential,huang2023belief,liu2024adaptive,liu2023adaptive}. However, real-world multi-view scenarios often involve challenges such as conflicting evidence across views and unreliable information from certain perspectives. Unfortunately, the traditional DST fusion framework is sensitive to evidence conflict and the order of fusion~\citep{huang2024trusted}, making it unable to robustly handle conflictive cases. 

Recent advances in TMVC \citep{xu2024reliable,huang2024trusted,yue2024evidential}, have made significant strides toward mitigating the challenges posed by conflicting evidence. Notably, RCML~\citep{RCML} introduces an conflict-aware fusion framework that reduces sensitivity to conflictive views, providing a more robust decision-making process. However, RCML overlooks an essential aspect of representation learning—the preservation of latent neighborhood structures among samples. This limitation can hinder the model’s ability to capture meaningful relationships within the data, which are crucial for resolving ambiguities and improving classification accuracy. On the other hand, TUNED~\citep{huang2024trusted} explicitly incorporates sample-level neighborhood structures by constructing adjacency matrices in Euclidean space and employing Graph Neural Networks (GNNs) for feature extraction. While effective in capturing structural relationships, \citep{huang2024trusted} reliance on explicit graph construction significantly increases its spatial and temporal complexity, making it less practical for large-scale applications. Furthermore, the explicit graph construction in TUNED assumes global consistency of the neighborhood structure, which may not always hold true in diverse multi-view scenarios. This motivates the need for a more flexible and computationally efficient approach to incorporate latent neighborhood structures. We argue that latent neighborhood structures can be implicitly considered without explicit graph construction. Instead of constructing a graph for the entire dataset, we propose to focus on the local structure and construct class-level prototypes. This approach simplifies computational complexity while retaining the essential geometric relationships required for robust representation learning. Building on this insight, we introduce a novel framework that integrates latent neighborhood structures into the multi-view classification process while maintaining computational efficiency.

Our method operates in two stages. First, we construct structure-aware prototypes that implicitly capture the local neighborhood relationships within mini-batches. These prototypes serve as anchors for class-level information, enabling the model to guide evidence fusion across views while preserving latent neighborhood structures. Second, we perform prototype-guided fine-grained evidence fusion, where the contribution of each view is adaptively weighted at class level. We found that the representations obtained by transforming the prototypes through the evidence extractor exhibit evidence-grounded properties and can effectively capture uncertainty. By combining structure-aware prototype learning with prototype-guided fusion, our framework effectively addresses the challenges posed by conflictive and unreliable views, offering a principled solution for TMVC. Our contributions are given as follows:

\begin{itemize}
    \item We introduce a structure-aware prototype learning mechanism that implicitly considers latent neighborhood structures without requiring explicit graph construction, significantly reducing computational complexity.
    \item We develop a prototype-guided fine-grained fusion (PFF) strategy that dynamically adjusts view contributions at class level, mitigating the impact of conflicting evidence.
    \item We demonstrate the effectiveness of our framework through extensive experiments, achieving competitive performance compared to prevalent TMVC methods, while maintaining computational efficiency.
\end{itemize}
\vspace{-0.8em}

\begin{figure*}
    \centering
    \includegraphics[width=1\linewidth]{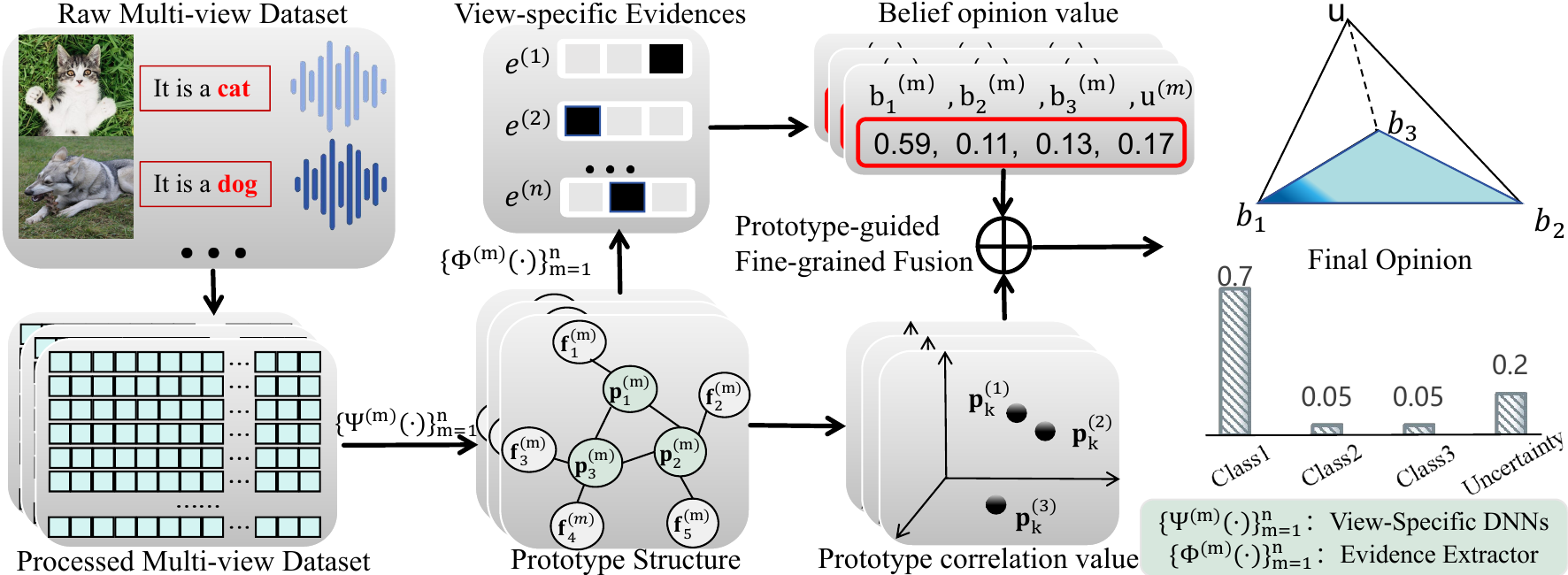}
    \vspace{-2em}
    \caption{Overview of the proposed framework. The process begins by transforming multi-view data into structured feature representations using view-specific deep neural networks. Evidence is then derived from both view-specific outputs and prototype-based embeddings that dynamically encode intra- and inter-view relations. These sources of evidence are subsequently integrated through Prototype-Guided Fine-Grained Fusion (PFF), yielding robust predictions along with quantified uncertainty.}
    \vspace{-1em}
    \label{fig2}
\end{figure*}

\section{Related Work}
\label{sec:related_work}
Multi-view learning leverages complementary information from diverse sources, but real-world data often contain noise, inconsistencies, or adversarial perturbations that induce inter-view conflicts. Most existing conflictive multi-view methods aim to eliminate such conflicts, typically via outlier detection that assesses cross-view consistency. These methods are mainly categorized into cluster-based~\citep{Huang_Ren_Pu_Huang_Xu_He_2023,zhang2023let,liu2024adaptive,marcos2013clustering,zhao2017consensus} and self-representation-based methods~\citep{wang2019adversarial,wen2023highly,hou2020fast}. The former clusters each view independently and compares affinity vectors for outlier detection, while the latter reconstructs instances using typical samples. Another line addresses partially aligned views~\citep{zhang2021late,wen2023unpaired}, evolving from weakly-paired maximum covariance analysis~\citep{lampert2010weakly} to differentiable Hungarian matching~\citep{huang2020partially} and robust contrastive learning~\citep{Yang_2021_CVPR,li2022positive,qin2024noisy}. These methods aim for consensus by suppressing conflicts but often fail in inherently conflicting scenarios like adversarial defense or uncertainty-aware tasks.

Trusted multi-view learning offers a promising alternative. Although deep networks achieve strong performance, they often lack reliable uncertainty estimates under noise or distribution shifts~\citep{wen2023deep,chen2024finecliper,hu2024recent}. Solutions include deterministic uncertainty~\citep{sensoy2018evidential}, Bayesian methods~\citep{gal2016dropout}, ensembles~\citep{lakshminarayanan2017simple}, and test-time augmentation~\citep{lyzhov2020greedy}. Among them, Evidential Deep Learning (EDL), based on subjective logic~\citep{josang2016subjective} and Dempster-Shafer theory~\citep{dempster1968generalization}, has excelled in multi-view classification~\citep{han2021trusted,han2022trusted,Liu_Yue_Chen_Denoeux_2022,liu2023safe,xu2024reliable,huang2024evidential,qiu2025l2,liu2024adaptive}, zero-shot learning~\citep{huang2024crest}, cross-modal retrieval~\citep{qin2022deep}, out-of-distribution (OOD) detection~\citep{qu2024hyper}, open-world multi-view classification~\citep{dong2025trusted}, video understanding~\citep{chen2024uncertainty} and action localization~\citep{Chen2023TAL,Gao2023TAL,ma2024beyond}. In conflictive multi-view learning, EDL has been used to quantify uncertainty over global proxy views~\citep{liu2023safe,Liu_Yue_Chen_Denoeux_2022,xu2024reliable} or detect local intra-view inconsistencies~\citep{yue2024evidential}. However, these methods rely on Dempster-Shafer fusion, which is sensitive to conflicting evidence and prone to unreliable results without well-designed losses~\citep{xiao2019multi,huang2023belief}. Moreover, most treat representation learning and uncertainty modeling separately, limiting adaptive conflict integration.

To overcome these issues, we propose a framework that employs EDL-based uncertainty modeling with class-level prototypes, enabling robust multi-view decision-making under conflict without depending on handcrafted similarity metrics or rigid fusion rules.

\section{Methodology}
\label{sec:methodology}
\subsection{Prototype-Guided Multi-view Learning}
\label{subsec:1}

To address the challenges posed by heterogeneous feature spaces and structural inconsistencies across views, we propose a prototype-guided multi-view learning framework. In multi-view data, modal discrepancies and inconsistent patterns often introduce uncertainty during the fusion process. Although EDL effectively models uncertainty within individual views, it is highly sensitive to conflicting evidence and often neglects the crucial role of local neighborhood structures. Traditional methods that incorporate neighborhood information typically rely on explicit graph modeling, leading to substantial computational overhead. To overcome these limitations, we leverage prototypes as anchors for class-level information and preserve latent neighborhood structures in a computation-friendly manner. By integrating prototypes with latent local neighborhood and discriminative constraints, our framework achieves robust, interpretable, and unified feature alignment across views, effectively mitigating the impact of conflicting or uncertain evidence.

\textbf{Prototype Construction.} \label{subsubsec:1}In multi-view learning, challenges arise due to structural inconsistencies across views, making direct comparison or fusion difficult. Traditional methods like \citep{RCML} often overlook the structural nuances within and between views, while approaches such as  \citep{huang2024trusted} utilize multi-view GNNs to explicitly model graph structures, incurring significant computational costs.

Our proposed method addresses these issues by projecting each view into a latent space $\mathcal{F}$ using view-specific deep neural networks (DNNs) as follows: 
\vspace{-0.6em}
\begin{equation}
\label{eq:1}
\vspace{-0.8em}
    \mathcal{F}^{(m)} = \Psi^{(m)}(X^{(m)}), \quad \mathcal{F}^{(m)} \in \mathbb{R}^{N \times d}
\end{equation}
where $\Psi^{(m)}(\cdot)$ is the feature encoder for view $m$, $N$ is the number of samples, and $d$ is the embedding dimension. 

Then, we perform average pooling on the features extracted from the samples to obtain class-level representations, which serve as the prototype.
\vspace{-0.6em}
\begin{equation}
    \mathbf{p}_k^{(m)} = \frac{1}{|\mathcal{N}_k^{(m)}|}\sum\limits_{i=1}^{|\mathcal{N}^{(m)}_k|}\mathbf{f}_i^{(m)}
    \vspace{-1em}
\end{equation}
where \(\mathcal{N}_k^{(m)}\) is the set of samples belonging to class \(k\) in view \(m\) with a sample size $|\mathcal{N}_k^{(m)}|$  and \(\mathbf{f}_i^{(m)}\) represents the feature embedding of the \(i\)-th sample in \(\mathcal{N}_k^{(m)}\).

Finally, a learnable evidence mapping module, denoted as $\Phi(\cdot)$, is employed to project the extracted features into non-negative evidence space.
\vspace{-0.6em}
\begin{equation}
    \mathcal{E}^{(m)} = \Phi(\mathcal{F}^{(m)}),\quad \mathcal{E}^{(m)}\in\mathbb{R}^{N\times K}
    \vspace{-1em}
\end{equation}

Similarly, the prototype is processed by the evidence extractor to obtain a representation aligned with the evidence dimension:
\begin{equation}
\mathbf{q}_k^{(m)} = \Phi(\mathbf{p}_k^{(m)}),
\vspace{-1em}
\end{equation}
where $\mathbf{q}_k^{(m)}$ is the embedding of prototype $\mathbf{p}_k^{(m)}$ with the same dimensionality as the evidence.

\begin{table}[t]
\centering
\resizebox{\linewidth}{!}{ 
\renewcommand{\arraystretch}{1} 
\setlength{\tabcolsep}{3pt} 
\begin{tabular}{@{}p{2.2cm}p{2.2cm}p{2.8cm}p{2.8cm}p{2.8cm}@{}}
\toprule
\textbf{Method} & \textbf{SA} & \textbf{FE} & \textbf{Training} & \textbf{Overall} \\ 
\midrule
RCML            & $\times$      & $\mathcal{O}(nd)$          & $\mathcal{O}(n^2d)$         & $\mathcal{O}(n^2d+nd)$         \\
TUNED           & $\checkmark$  & $\mathcal{O}(nd)$          & $\mathcal{O}(n^2d + n^3Ld)$ & $\mathcal{O}(n^2d + n^3Ld)$ \\
Ours            & $\checkmark$  & $\mathcal{O}(nKd)$         & $\mathcal{O}(n^2d)$               & $\mathcal{O}(n^2d+nKd)$          \\ 
\bottomrule
\end{tabular}
}
\vspace{-1em}
\caption{Comparison of computational complexity}
\begin{tablenotes}
\scriptsize
\item \textbf{Notes:} \textbf{FE} denotes ``Feature Extraction". \textbf{SA} indicates whether neighbor structure is considered, $L$ is the number of views, and $K$ is the number of classes.
\end{tablenotes}
\label{tab:complexity}
\vspace{-1.5em}
\end{table}

By using prototypes as highly abstract representations of classes, one can effectively capture class features across different views and encode neighborhood information in a computationally efficient manner. As shown in \textbf{Table~\ref{tab:complexity}}, RCML~\citep{RCML} relies on pairwise similarity computations in the embedding space to mitigate conflicts, resulting in a computational complexity of $\mathcal{O}(n^2d)$. TUNED~\citep{huang2024trusted} further increases the computational cost by constructing a weighted graph ($\mathcal{O}(n^2d)$) and applying GNNs for global neighbor feature extraction ($\mathcal{O}(n^3Ld)$), leading to an overall complexity of $\mathcal{O}(n^2d + n^3Ld)$. In contrast, our method eliminates the need for pairwise computations and explicit graph construction. By leveraging class prototypes and view-specific embeddings, we reduce feature extraction complexity to $\mathcal{O}(nKd)$, achieving linear scalability with respect to $n$. This makes our approach significantly more efficient and scalable for large datasets, while maintaining competitive performance.

To ensure that prototypes serve as effective guidance during training, we designed three loss functions: \ding{182} Contrastive Prototype Learning Loss, which leverages positive and negative partitioning to enhance the discriminative capability of prototype features; \ding{183} Label Alignment Loss, which encourages prototypes to align with their corresponding class labels and remain well-separated across views; \ding{184} Neighbor Structure Alignment Loss, which encourages prototypes to capture the underlying neighborhood structure.

\textbf{Contrastive Prototype Learning.}
To enhance feature discriminability and encourage features of the same class to cluster around their corresponding prototypes, we introduce a contrastive prototype learning loss. This loss drives features toward their class-specific prototype while pushing them away from prototypes of other classes, thereby promoting compact prototype-centered clusters. For a feature \( \mathbf{f}^{(m)} \), let \( \mathbf{p}^{(m)}_{+} \) denote the positive prototype (same class) and \( \mathbf{p}^{(m)}_{-} \) denote the negative prototype (different class). The loss is defined as:
\vspace{-0.6em}
\begin{equation}
\label{eq:3}
\begin{aligned}
     &\mathcal{L}_{\text{c}} = \frac{1}{K}\frac{1}{\lvert R \rvert}\sum_{k=1}^{K}\sum_{j=1}^{|R|} \\
     &\max \left( D_{KL}( \mathbf{p}_{+}^{(m)},\mathbf{f}_{j}^{(m)}) - D_{KL}( \mathbf{p}_{-}^{(m)},\mathbf{f}_{j}^{(m)}) + \lambda, 0 \right),
\end{aligned}
\vspace{-0.3em}
\end{equation}
where $D_{\text{KL}}$ denotes the KL divergence, $\lambda$ is a margin that enforces the positive similarity to exceed the negative similarity by at least $\lambda$. $R$ is the KNN feature set associated with view $m$ and class $k$, and $|R|$ indicates the number of neighbor samples selected for each prototype.

\textbf{Label Alignment Loss.}
We introduce a label alignment loss based on KL divergence to ensure that prototypes from each perspective are both representative and well-separated. On one hand, we encourage the prototypes to align with their corresponding class labels and maintain consistency across different views, thereby alleviating discrepancies among different modalities. On the other hand, this learning strategy enhances inter-class discriminability and prevents prototype collapse.
\vspace{-0.4em}
\begin{equation}
\begin{aligned}
\label{eq:4}
    &\mathcal{L}_{t} = \frac{1}{K}\sum\limits_{m=1}^{M}\sum_{k=1}^{K}D_{KL}( \mathbf{q}_{k}^{(m)},\mathbf{y}_{k})\\
    &+\xi\frac{1}{K(K-1)}\sum\limits_{m=1}^{M}\sum\limits_{k_1=1}^{K}\sum\limits_{k_2=1,k_2\neq k_1}^{K}\frac{\mathbf{q}_{k_1}^{(m)\top}\mathbf{q}_{k_2}^{(m)}}{\|\mathbf{q}_{k_1}^{(m)}\|\|\mathbf{q}_{k_2}^{(m)}\|},
\end{aligned}
\end{equation}
where $K$ is the number of classes and $\mathbf{y_k}$ is the one-hot encoded vector for class $k$. $\xi$ refers to the regularization parameter.

\textbf{Neighbor Structure Alignment.}
Traditional methods often overlook local neighborhood structures, leading to suboptimal performance. To better capture essential relational information, we introduce a neighbor-alignment loss that ensures consistency between each prototype and its local neighborhood. Specifically, using KNN, the top-$|R|$ nearest samples are selected to align each prototype with its most relevant neighbors as follows:

\vspace{-0.4em}
\begin{equation}
    \mathcal{L}_{n} = \frac{1}{K}\cdot\frac{1}{\lvert R \rvert }\sum\limits_{m=1}^{M}\sum_{k = 1}^{K}\sum_{j=1}^{|R|}D_{KL}( \mathbf{p}_{k}^{(m)},\mathbf{f}_{j}^{(m)})
\vspace{-0.4em}
\end{equation}
It is worth noting that $K$ has been previously defined as the number of classes in the dataset; therefore, we adopt $|R|$ to indicate the number of samples selected by KNN. The $|R|$ here is consistent with the $|R|$ used in Contrastive Prototype Learning.

\subsection{Structure-Aware Conflictive Multi-view Learning}

\textbf{View-specific Evidential Deep Learning.}
Most multi-view methods rely on softmax, causing overconfidence and ignoring uncertainty. We address this by integrating EDL~\citep{sensoy2018evidential}, which uses evidence theory to interpret data cues for class decisions. For the sample set in each view, the evidential neural network $\Psi^{(m)}(\cdot)$ extract evidence $\{\mathbf{\mathcal{E}^{(m)}}\}$. In a $K$-class setting, each view produces the corresponding opinion triplet $\omega = (b, u)$, where belief mass $b = (b_1, \dots, b_K)^\top$ quantifies class support, uncertainty mass $u$ reflects lack of evidence. Following subjective logic, $b$ and $u$ follow the rule below:
\vspace{-0.4em}
\begin{equation}
    \sum_{k=1}^{K} b_k + u = 1, \quad \forall k \in [1, \ldots, K],
\end{equation}
where $b_k$ $\geq$ 0, u $\geq$ 0. And $\mathcal{S}_K$ is the K-dimensional unit simplex, defined as:
\vspace{-0.4em}
\begin{equation}
    \mathcal{S}_K = \left\{ \mathbf{S} \, \middle| \, \sum_{k=1}^{K} s_k = 1 \text{ and } 0 \leq s_1, \ldots, s_k \leq 1 \right\}.
\end{equation}

With Dirichlet parameters are set as \( \alpha = e + 1 \) to ensure non-sparsity. Denoting \( S = \sum_{k=1}^{K} (\alpha_k) \) be the Dirichlet strength, the corresponding belief mass and uncertainty for view $m$ can be computed as:
\vspace{-0.5em}
\begin{equation}
    b_k^{(m)} = \frac{e_k^{(m)}}{S} = \frac{\alpha_k^{(m)} - 1}{S}, \quad u^{(m)} = \frac{K}{S},
\end{equation}
and the class probability is \( c_k = \frac{\alpha_k}{S} \). Through the view-specific evidential learning stage, we obtain the view-specific opinion and the corresponding class distribution.

\textbf{EDL Loss.}
The EDL loss function consists of two parts:

\textbf{(1) Classification Loss}:  
The classification loss aligns the model's predicted beliefs with the true class labels.
\vspace{-0.6em}
\begin{equation}
\mathcal{L}_{\text{ace}}(\alpha_n) = \sum_{j=1}^K y_{nj} \left( \psi(S_n) - \psi(\alpha_{nj}) \right),
\vspace{-1em}
\end{equation}

where \(S_n = \sum_{j=1}^K \alpha_{nj}\) is the total evidence, and \(\psi(\cdot)\) is the Digamma function.

\textbf{(2) Divergence Loss}:  
This term regularizes the model by penalizing deviations between the predicted beliefs and the prior distribution, expressed as:
\vspace{-0.4em}
\begin{equation}
\begin{split}
\mathcal{L}_{\text{D}}(\alpha_n) &= \log \frac{\Gamma(K)}{\prod_{k=1}^K \Gamma(\alpha_{nk})} \\
&+ \sum_{k=1}^K (\alpha_{nk} - 1) \left( \psi(\alpha_{nk}) - \psi(S_n) \right),
\vspace{-1em}
\end{split}
\end{equation}
where \(\Gamma(\cdot)\) is the Gamma function.

The final \textbf{EDL loss} combines these components:
\begin{equation}
\vspace{-0.3em}
\mathcal{L}_{\text{EDL}} = \mathcal{L}_{\text{ace}}(\alpha_n) + \mu_t \mathcal{L}_{\text{D}}(\alpha_n),
\end{equation}
where $\mu_t$ = min (1, $\frac{t}{\mathcal{T}}$ ) $\in$ [0, 1] is the annealing coefficient, with $t$ representing the current training epoch and $\mathcal{T}$ the total number of annealing steps.

To sum up, the overall loss function for a specific instance can be calculated as:
\begin{equation}
\vspace{-0.3em}
    \mathcal{L} = \mathcal{L}_{\text{EDL}} + \alpha \mathcal{L}_{\text{t}} + \beta \mathcal{L}_{\text{n}} + \gamma \mathcal{L}_{\text{c}}
\end{equation}
where $\alpha$, $\beta$, and $\gamma$ are the weighting coefficients that regulate the relative contributions of each respective loss term.

\subsection{Prototype-Guided Fine-Grained Fusion}

Existing multi-view fusion methods typically adopt global weighting schemes that treat all classes uniformly, overlooking class-level diversity~\citep{dong2025trusted,han2021trusted,liu2023safe}. In practice, different classes may rely unequally on specific views or feature subsets—some are better represented in certain views, while others depend on distinct view-specific characteristics. To capture this heterogeneity, view weights should vary across classes rather than remain fixed globally.

To this end, we propose Prototype-Guided Fine-Grained Fusion (PFF) mechanism, which leverages prototypes to guide class-level multi-view fusion. Prototypes, as compact representations of class characteristics, encode local neighborhood structures and facilitate fine-grained feature analysis. By incorporating prototype guidance, PFF adaptively models class-dependent view importance, enabling more precise conflict detection and significantly enhancing the robustness of multi-view evidence fusion.

\textbf{Belief Opinion Value.} 
During the fusion process, we hope to preserve the individuality of each sample. In this aim, we evaluate the contribution of each view to the overall class evidence through a confidence quality measure, which is computed by normalizing the evidence based on the Dirichlet intensity. If the evidence extracted from a sample in a particular view exhibits a higher belief value for a given class, it indicates that this view holds stronger confidence in that prediction. Accordingly, the corresponding weight for that view is assigned a higher value. For view m and class $k$, it is calculated as:
\vspace{-0.3em}
\begin{equation}
    \mathcal{B}_k^{(m)}  =  \frac{e_{k}^{(m)}}{S^{(m)}}
\vspace{-0.8em}
\end{equation}
where $e_{k}^{(m)}$ represents the component of evidence $\mathbf{e}^{(m)}$ in the $k$-th class for view $m$ and $S^{(m)}$ is the Dirichlet strength for view $m$.

\begin{figure}
    \centering
    \includegraphics[width=1\linewidth]{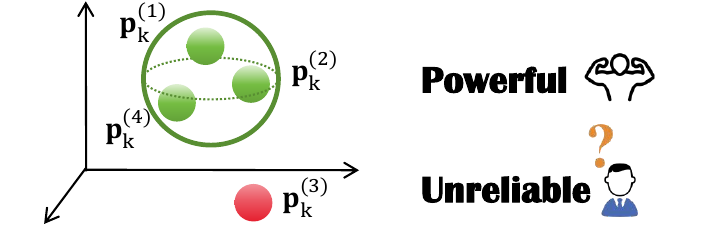}
    \vspace{-1.6em}
    \caption{Effect of prototype correlation values}
    \label{fig:correlation}
    \vspace{-1em}
\end{figure}

\textbf{Prototype Correlation Value.}
To better identify and suppress unreliable views, we introduce a prototype correlation value to measure cross-view structural alignment. As shown in \textbf{Figure~\ref{fig:correlation}}, views whose prototypes are highly consistent with others are considered more reliable, while those with low agreement are less trustworthy. Leveraging this, PFF adaptively weights views at the class level, emphasizing reliable information and reducing the impact of conflicting or noisy evidence. The correlation value is formally defined as follows:
\begin{equation}
\mathcal{C}_k^{(m)} = \frac{\sum\limits_{j=1,j \neq m}^{M} \exp(-D_{\text{KL}}(\mathbf{p}_k^{(m)} , \mathbf{p}_k^{(j)}))}{\sum\limits_{i=1}^{M} \sum\limits_{j=1,j \neq i}^{M} \exp(-D_{\text{KL}}(\mathbf{p}_k^{(i)}, \mathbf{p}_k^{(j)}))}.
\vspace{-0.6em}
\end{equation}

\textbf{Multi-view Evidence Fusion.}
Although prototype relevance can partially expose view-level conflicts, its reliability remains insufficiently established. By introducing the Label Alignment Loss to align prototypes with the evidence extractor, both are embedded into a shared evidence space, which justifies extending uncertainty modeling from evidence to prototypes. 

To further examine this hypothesis, we partition classes into uncertainty intervals and compute, for each interval, the prediction correctness of samples whose predicted class falls within it (averaged across views; fusion excluded). As shown in \textbf{Figure~\ref{fig:uncertainty}}, prototype-derived evidence uncertainty exhibits an approximately inverse relationship with correctness. This observation indicates that the representations extracted from prototypes via the evidence extractor possess evidence-grounded characteristics and can thus serve to assess prototype reliability. Consequently, we subtract the uncertainty $u_k^{(m)}$ of each class prototype from the final value as follows:

\begin{figure}
    \centering
    \includegraphics[width=1\linewidth]{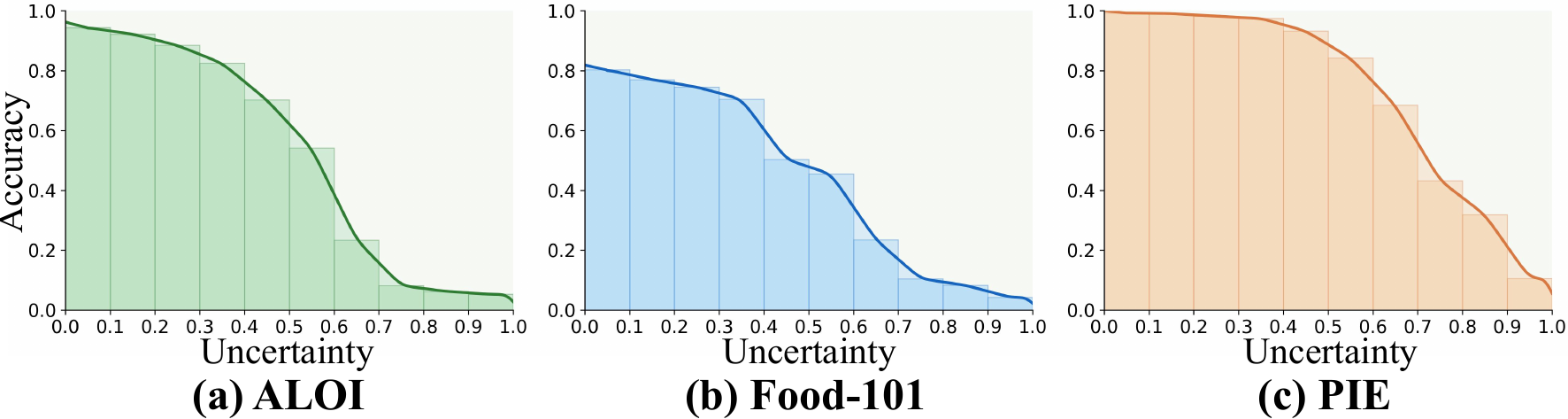}
    \vspace{-1.6em}
    \caption{Illustration of the relationship between prototype-derived evidence uncertainty and prediction correctness on three datasets.}
    \label{fig:uncertainty}
    \vspace{-1.8em}
\end{figure}

\begin{equation}
    v_{k}^{(m)} = \mathcal{B}_{k}^{(m)} + \mathcal{C}_{k}^{(m)} - u_k^{(m)}
    \vspace{-0.4em}
\end{equation}

The weights for view $m$ and class $k$ are derived through normalization as follows:
\begin{equation}
\vspace{-0.8em}
    w_{k}^{(m)} = \frac{v_k^{(m)}}{\sum\limits_{i=1}^{M}v_k^{(i)}}
    \vspace{-0.2em}
\end{equation}

For each sample, the fused evidence is computed as a weighted sum:

\begin{equation}
    e_k = \frac{1}{M}\sum_{m=1}^M (w^{(m)}_k\cdot e^{(m)}_k)
\vspace{-0.8em}
\end{equation}
With the final fused evidence $\mathbf{e} = [e_1, e_2, \dots,e_K]$.

\section{Experiment}
\label{sec:experiment}
\subsection{Experimental setup}

\textbf{Datasets.}
\textbf{Table~\ref{tab1}} provides an overview of the datasets used in our experiments. We evaluated our method on six multi-view datasets spanning image recognition and sentiment analysis, including PIE, HandWritten, ALOI, NUS-WIDE-OBJECT, MOSI, and Food-101. Each dataset contains multiple feature views or modalities representing diverse data characteristics. Detailed dataset descriptions are provided in the \textbf{Supplementary Material}.

\begin{table}[ht]
    \centering
    \small
    \begin{tabular}{cccc}
    \toprule
       Dataset  &  Size  &  Classes  &  Dimensionality\\
    \midrule
       PIE  &  680  &  68  &  484\textbar256\textbar279 \\
       HandWritten  &  2000  &  10  &  240\textbar76\textbar216\textbar47\textbar64\textbar6 \\
       ALOI  &  10800  &  100  &  77\textbar13\textbar64\textbar125\\
       NUS & 30000 &  31  & 129\textbar74\textbar145\textbar226\textbar65 \\
       MOSI  &  1284  & 3  &  768\textbar5\textbar20\\
       Food-101 & 87696 & 101 & 512\textbar512\textbar768 \\
    \bottomrule
    \end{tabular}
    \vspace{-0.8em}
    \caption{Overview of datasets used in experiments}
    \label{tab1}
    \vspace{-1em}
\end{table}

\textbf{Compared Methods.}
We compared our approach with nine representative methods. Among them, EDL~\citep{sensoy2018evidential} is a single-view baseline, while TMC~\citep{han2021trusted}, TMDL-OA~\citep{Liu_Yue_Chen_Denoeux_2022}, RCML~\citep{xu2024reliable}, and RMVC~\citep{yue2024evidential} are TMVC methods addressing uncertainty and conflict; TUNED~\citep{huang2024trusted} performs multi-view evidence fusion via Selective Markov random field (S-MRF). Details are in the \textbf{Supplementary Material}.

\textbf{Implementation Details}.
The Adam optimizer~\citep{AE2} is used to train the network, where l2-norm regularization is set to 1$e^{-5}$.  Throughout our experiments, the loss function weights were fixed to the optimal values obtained through preliminary tuning. For all datasets, 20\% samples are used as test set. The conflictive datasets were constructed by us based on these datasets, following the methodology outlined in~\citep{xu2024reliable}. We run 10 times for each method to report the mean values and standard deviations. The prototype's dimensions are set to 1024. The model is implemented by PyTorch on one NVIDIA A100 with GPU of 40GB memory.

\subsection{Comparison with State-of-the-Arts}
\textbf{Table~\ref{tab2}} and \textbf{Table~\ref{tab3}} show the performance on normal and conflictive test sets. We observe that: (1) On normal test sets, our method outperforms all baselines on all datasets except HandWritten. (2) On conflictive datasets, our method demonstrates considerable advantages. Except for the ALOI dataset, it ranks second only to TUNED, achieving the highest accuracy across the remaining five datasets. Notably, it outperforms RCML, by 12.61\% on the MOSI dataset.

\begin{table*}[ht]
\centering
\small
\begin{tabular*}{1.00\textwidth}{@{\extracolsep{\fill}}ccccccccc}
\toprule
Method & PIE & HandWritten & ALOI & NUS & MOSI & Food-101\\
\midrule
\rowcolor{cyan!10} EDL & 86.25±0.89 & 96.90±0.16 & 37.87±5.13 & 22.33±0.64 & 51.35±1.77 & 55.68±1.00\\
\rowcolor{cyan!10} DCCAE & 81.96±1.04 & 95.45±0.35  & 87.20±0.14 & 35.75±0.48 &50.47±0.69 & 64.17±0.77\\
\rowcolor{cyan!10} CPM-Nets & 88.53±1.23 & 94.55±1.36  & 49.13±6.19 & 35.37±1.05 &51.29±0.51 & 66.31±0.62\\
\rowcolor{cyan!10} DUA-Nets & 90.56±0.47 & 98.10±0.32 & 83.09±2.15 & 33.98±0.34 &56.67±0.14 & 67.87±1.01\\
\rowcolor{cyan!10} TMC & 91.85±0.23 & 98.51±0.13 & 79.31±0.31 & 35.18±1.55 & 61.14±0.97 & 69.01±0.53 \\
\rowcolor{cyan!10} TMDL-OA & 92.33±0.36 & \underline{99.25±0.45} & 65.20±0.18 & 34.39±0.44 &63.24±0.66 & 68.56±0.42\\
\rowcolor{cyan!10} RCML & 94.71±0.02 & \textbf{99.40±0.00} & 69.29± 0.11 & 34.04 ± 0.27 &67.20±0.28 &70.17±0.52 \\
\rowcolor{cyan!10} RMVC & 91.18±0.24 & 98.51±0.04 & 87.38±0.67 & 34.68±0.32 &66.51±0.49 & 68.47±0.51\\
\rowcolor{cyan!10} TUNED & \underline{96.83±0.01} & 99.20±0.23  & \underline{88.93±0.35} & \underline{37.46±0.25} &\underline{70.39±0.19} & \underline{72.44±0.11}\\
\rowcolor{green!10} Ours  &  \textbf{98.53±0.02}  &  99.00±0.25  & \textbf{91.16±0.73} & \textbf{38.20±0.04} &\textbf{72.89±0.21} &\textbf{74.49±0.07}\\
\rowcolor{yellow!10} $\bigtriangleup \%$  & 1.76 & -0.40 & 2.51 & 1.98 & 3.55 & 2.83 \\
\bottomrule
\end{tabular*}
\vspace{-1em}
\caption{Accuracy (\%) on normal test sets. \textbf{Bold} denotes the best, while \underline{underline} means the suboptimal. }
\label{tab2} 
\end{table*}

\begin{table*}[ht]
\centering
\vspace{-2mm}

\small
\begin{tabular*}{1.00\textwidth}{@{\extracolsep{\fill}}cccccccc}
\toprule
Method & PIE & HandWritten & ALOI & NUS & MOSI &Food-101\\
\midrule
\rowcolor{cyan!10} EDL & 21.76±0.67 & 57.25±0.28 & 25.05±3.10 & 18.07±0.28 & 37.49±2.96 &44.25±1.59\\
\rowcolor{cyan!10} DCCAE & 26.89±1.10 & 82.85±0.38 & 75.12±0.43 & 32.12±0.52 & 41.58±0.61 &43.02±2.56\\
\rowcolor{cyan!10} CPM-Nets & 53.19±1.17 & 83.34±1.07 & 36.29±5.02 & 29.20±0.81 & 43.25±0.27 & 49.55±0.58\\
\rowcolor{cyan!10} DUA-Nets & 56.45±1.75 & 87.16±0.34 & 69.07±2.50 & 31.82±0.43 & 45.91±0.44 & 49.23±0.53\\
\rowcolor{cyan!10} TMC & 61.65±1.03 & 92.76±0.15 & 76.68±0.32 & \underline{33.76±2.16} &  47.77±0.19 & 51.45±0.68\\
\rowcolor{cyan!10} TMDL-OA & 68.16±0.34 & 93.05±0.05 & 62.90±0.12 & 32.44±0.26 & 49.76±0.47 &50.65±0.97\\
\rowcolor{cyan!10} RCML & 84.00±0.14 & 94.40±0.05 & 64.91 ± 0.20 & 31.19 ± 0.22 & \underline{58.12±0.51} & 62.56±0.71\\
\rowcolor{cyan!10} RMVC & 76.47±3.43 & \underline{94.75±0.75}  & 52.67±1.97 & 24.62±3.19 & 51.17±0.39 & 60.44±0.31\\
\rowcolor{cyan!10} TUNED & \underline{86.02±0.19} & 96.75± 0.55  & \textbf{88.49±0.29} & 34.09±0.14 & 55.25±0.33 & \underline{66.07±0.77}\\
\rowcolor{green!10} Ours & \textbf{86.76±0.73} & \textbf{97.50±0.25}  & \underline{85.46±0.17} & \textbf{34.13±0.12} & \textbf{65.45±0.19} & \textbf{68.31±0.11}\\
\rowcolor{yellow!10} $\bigtriangleup \%$ & 0.86 & 0.78 & -3.42& 0.12 & 12.61 & 3.39\\
\bottomrule
\end{tabular*}
\vspace{-1em}
\caption{Accuracy (\%) on conflictive test sets. \textbf{Bold} denotes the best, while \underline{underline} means the suboptimal.}
\label{tab3}
\vspace{-1em}
\end{table*}

\subsection{Ablation Study}
\begin{table}[ht]
\centering
\vspace{-0.8em}
\small
\begin{tabular}{@{\extracolsep{\fill}}lcccc}
\toprule
\multirow{2}{*}{Method} & \multicolumn{2}{c}{HandWritten} & \multicolumn{2}{c}{PIE} \\
\cmidrule(r){2-3} \cmidrule(r){4-5}
 & Normal & Conflict & Normal & Conflict \\
\midrule
w/o $\mathcal{L}_{\text{t}}$ & 96.75 & 95.75 & 96.43 & 85.75 \\
w/o $\mathcal{L}_{\text{n}}$ & 97.25 & 95.50 & 96.89 & 86.17 \\
w/o $\mathcal{L}_{\text{c}}$ & 96.75 & 95.50 & 95.92 & 85.58 \\
w/o $\mathcal{B}_k^{(m)}$    & 97.00 & 95.75 & 96.43 & 85.92 \\
w/o $\mathcal{C}_k^{(m)}$    & 97.25 & 95.25 & 96.89 & 85.75 \\
w/o $u_k^{(m)}$              & 97.25 & 96.00 & 96.57 & 86.09 \\
w/ Average                     & 97.50 & 96.00 & 96.71 & 86.44 \\
w/ DST                         & 98.25 & 97.25 & 96.44 & 86.17\\
w/ S-MRF                       & 98.75 & 97.25 & 96.83 & 86.22\\
w/ PFF & \textbf{99.00} & \textbf{97.50} & \textbf{98.53} & \textbf{86.76}\\
\bottomrule
\end{tabular}
\vspace{-1em}
\caption{Ablation study on HandWritten and PIE datasets under Normal and Conflict settings.}
\vspace{-1em}
\label{tab4}
\end{table}

As shown in \textbf{Table~\ref{tab4}}, each component of our method contributes to overall performance, as removing any loss term or PFF fusion component leads to degradation, particularly under view conflict. Specifically, excluding the label alignment ($\mathcal{L}_{\text{t}}$), neighbor alignment ($\mathcal{L}_{\text{n}}$), or contrastive loss ($\mathcal{L}_{\text{c}}$) slightly reduces accuracy on the PIE dataset. Moreover, our PFF fusion consistently outperforms other strategies such as arithmetic mean, Dempster, and S-MRF, confirming the effectiveness and robustness of all components, especially in conflictive scenarios.

\subsection{Qualitative Result}

\begin{figure}[ht]
    \centering
    \includegraphics[width=1\linewidth]{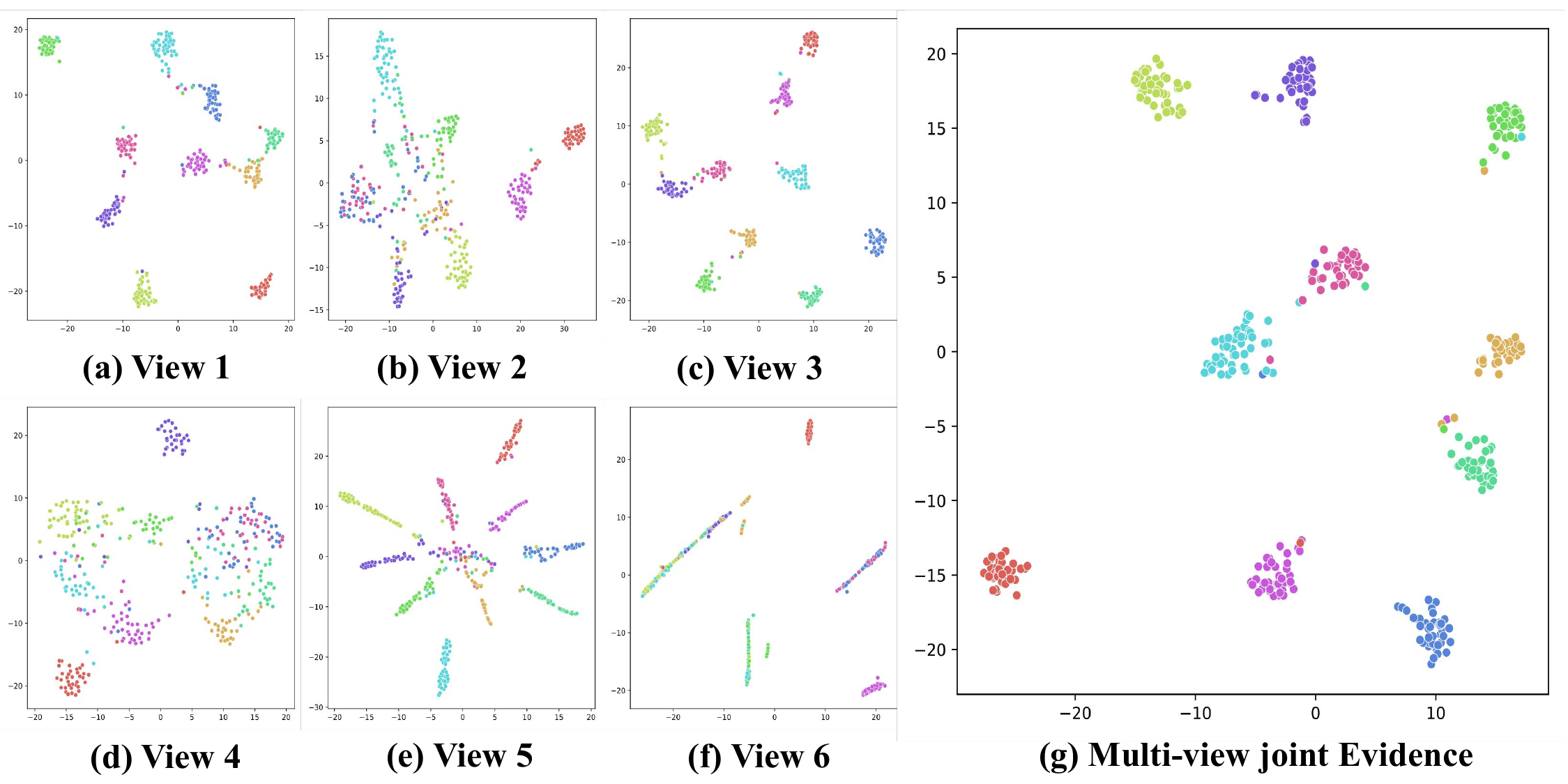}
    \vspace{-2em}
    \caption{Visualization of evidence distributions across six views and the fused joint evidence.}
    \label{tsne}
    \vspace{-0.8em}
\end{figure}

\textbf{T-SNE Visualization.}
\textbf{Figure~\ref{tsne}} visualizes evidences from three views and their fusion on HandWritten. Results show our Uncertainty-aware Evidence Fusion effectively captures inter-view relationships, producing a fused representation with clearer class boundaries and stronger discrimination.

\begin{figure*}[ht]
    \centering
    \includegraphics[width=1\linewidth]{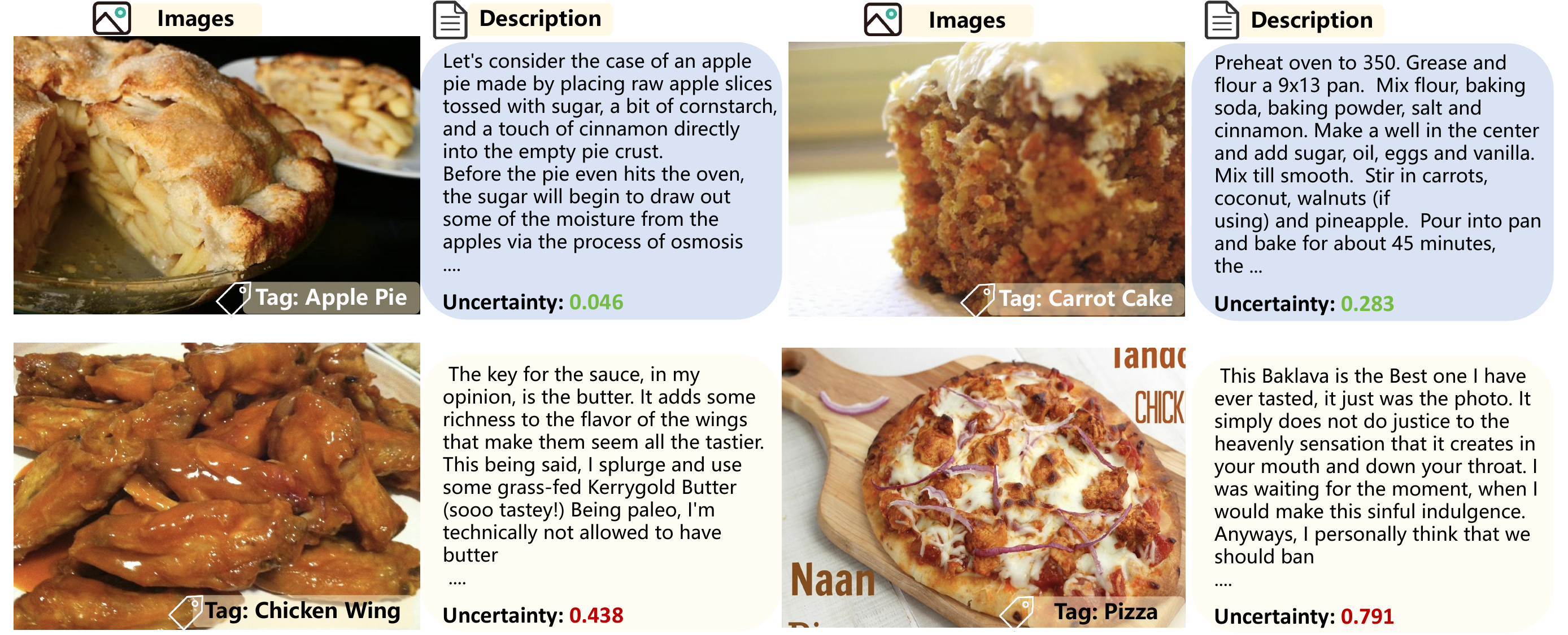}
    \vspace{-2.2em}
    \caption{Cases studies: visualization of uncertainty predictions on the Food-101 test set.}
    \label{fig4}
    \vspace{-0.4em}
\end{figure*}
\textbf{Case Study.}
The case studies in \textbf{Figure~\ref{fig4}} illustrate how our method effectively captures uncertainty in different Food-101 test samples. For the apple pie, with an uncertainty score of 0.046, the model shows high confidence, likely due to clear and consistent visual and textual features like raw apple slices, sugar, and cinnamon. In contrast, the carrot cake has a moderate uncertainty of 0.283, possibly due to overlapping features with other baked goods. The chicken wing shows higher uncertainty 0.438, potentially due to varied preparation styles and ingredients like butter-based sauces. The pizza shows the highest uncertainty 0.791, as the description is about a personal taste experience with baklava, conflicting with the pizza image—highlighting the model’s ability to detect such discrepancies.

\subsection{Further Analysis}

\textbf{Robust to Conflict.} 
Traditional evidence fusion frameworks are highly sensitive to the order of fusion and potential noise~\citep{huang2024trusted}. To evaluate robustness under view conflicts, we compare our method with TMDL-OA, RCML, and TUNED on the Handwritten dataset with varying conflict view combinations (Table~\ref{tab5}). Noise is added to simulate partial or severe corruption. While all methods show performance degradation as conflicts increase, our approach maintains greater stability, particularly in challenging cases like (2,5) and (0,1,2,3,4), where other methods experience significant drops.

Performance variance across combinations reflects view-dependent vulnerability. For example, corrupting view 4 severely impacts RCML and TMDL-OA but not ours, demonstrating the benefit of prototype-based alignment and uncertainty-aware weighting. Broader degradation in (0,2) or (0,2,4) suggests strong coupling among those views. Overall, our method better suppresses unreliable views and maintains consistency under conflict.

\begin{figure}[ht]
    \centering
    \vspace{-1em}
    \includegraphics[width=1\linewidth]{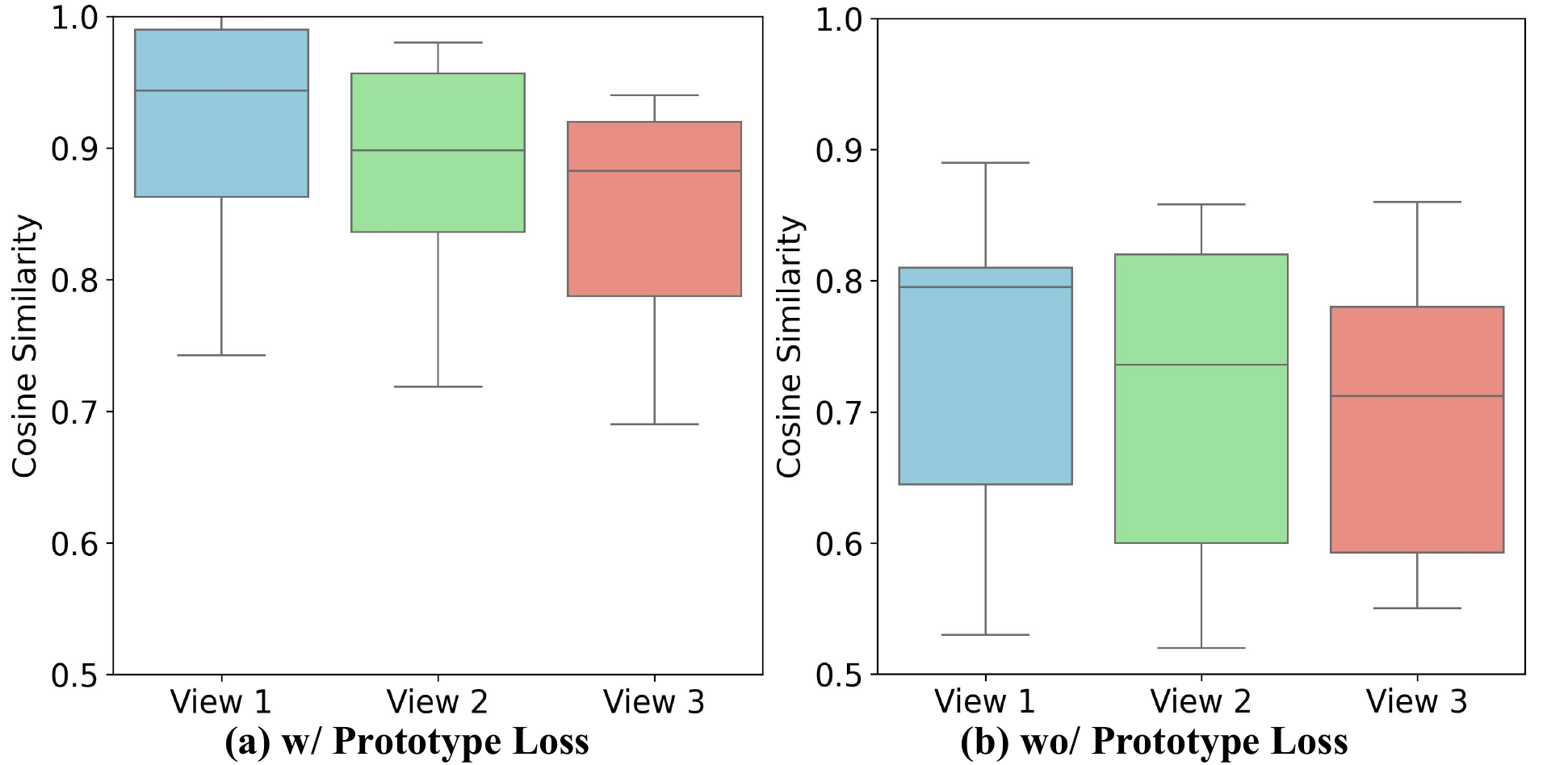}
    \vspace{-2em}
    \caption{The cosine similarity between features and prototypes within three views of the PIE dataset, comparing (a) trained with prototype loss $\mathcal{L}_\text{c}$ and (b) without.}
    \label{fig5}
\end{figure}

\begin{table}[ht]
    \centering
    \vspace{-0.8em}
    \small
    \begin{tabular}{ccccc}
    \toprule
    Conflict view(s) & Ours & TUNED & RCML & TMDL-OA \\
    \midrule
    0  &  96.25  &  \textbf{97.50} & 97.25 & 97.00 \\
    2  &  96.25  &  95.75 & 95.75 & \textbf{96.75} \\
    4  &  \textbf{97.50}  &  94.75 & 79.50 & 82.50 \\
    0,2  &  96.00 & \textbf{98.25} & 95.00 & 95.00 \\
    1,3  &  \textbf{97.00}  & 92.75 & 93.50 & 91.50 \\
    2,5  &  \textbf{95.50}  & 95.00 & 87.00 & 83.50 \\
    0,2,4 & 89.75 & \textbf{94.50} & 87.75  & 89.00 \\
    1,3,5 & \textbf{94.25} & 92.00 & 88.25 & 86.25 \\
    0,1,3,5 & \textbf{93.25} & 93.00 & 92.25 & 89.00 \\
    0,1,2,3,4 & \textbf{93.50} & 93.25 & 87.75 & 87.75 \\
    \bottomrule
    \end{tabular}
    \vspace{-1em}
    \caption{Accuracy (\%) of EDL-based methods with varying conflict views on the HandWritten dataset. Methods using the DST combination rule begin fusion from the first view, making them sensitive to the order of conflicting views.}
    \vspace{-1.4em}
    \label{tab5}
\end{table}

\textbf{Impact of Contrastive Prototype Loss $\mathcal{L}_\text{c}$.}
\textbf{Figure~\ref{fig5}} shows the effect of prototype loss on intra-view feature-prototype similarity on the PIE dataset. Comparing distributions with and without the loss reveals that incorporating it yields higher and more consistent alignment across views, thereby improving robustness and representation quality.


\textbf{Computational Time.}
We compare the average per-epoch training time and FLOPs of RCML, TUNED, and our method on PIE, HandWritten and ALOI datasets. Our method achieves similar efficiency to RCML and is notably faster than TUNED, consistent with the theoretical analysis. Details are provided in the \textbf{Supplementary Material}.

\begin{figure}[!t]
    \centering
    \includegraphics[width=1\linewidth]{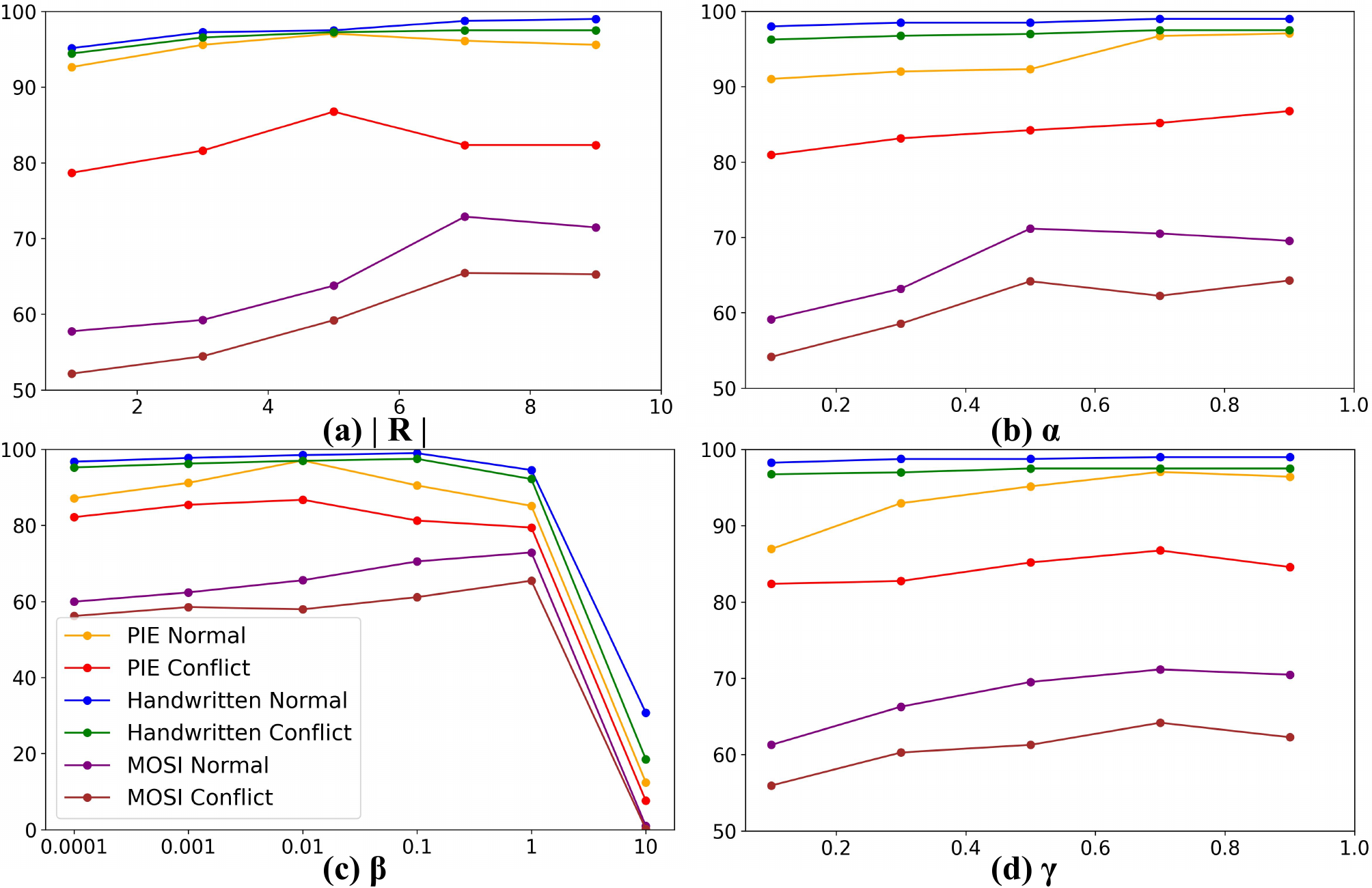}
    \vspace{-2em}
    \caption{Sensitivity analysis of model performance on various datasets with respect to key parameters under normal and conflictive conditions.}
    \label{fig10}
    \vspace{-1.5em}
\end{figure}

\textbf{Parameters Analysis.}
We evaluated the sensitivity of key hyperparameters, including the normalization parameter $\xi$, the KNN feature size $|R|$, and the loss weights $\alpha$, $\beta$, and $\gamma$, under both normal and conflictive conditions. For $\xi$, we tested values in ${1, 10^1, 10^2}$ and found its optimal choice remains largely empirical, so detailed discussion is omitted. \textbf{Figure~\ref{fig10}} shows the sensitivity of $|R|$, $\alpha$, $\beta$, and $\gamma$ on the PIE, Handwritten, and MOSI datasets under both settings. Detailed analyses are provided in the \textbf{Supplementary Material}.

\section{Conclusion}
\label{sec:conclusion}
In this study, we proposed Prototype-Guided Multi-view Learning based on a novel framework to address the challenges of multi-view classification, particularly in scenarios with conflicting views and high uncertainty. To better handle uncertainty and improve robustness, we incorporated EDL into the view-specific learning process, which probabilistically models class distributions and mitigates the impact of conflicting data. Based on prototype-guided fine-grained fusion strategy, we dynamically adjust the contribution of each view at the category level, thereby mitigating the impact of conflicting evidence. Extensive experiments are carried out to confirm the effectiveness of our method.

{
    \small
    \bibliographystyle{ieeenat_fullname}
    \bibliography{main}
}

\clearpage
\setcounter{page}{1}
\maketitlesupplementary

%
%
%

\section{Experiment setup}
\subsection{Dataset}
As for \textbf{normal datasets}, $\textbf{PIE}$\footnote{http://www.cs.cmu.edu/afs/cs/project/PIE/MultiPie/Multi-Pie/Home.html} contains 680 samples across 68 classes, represented by three feature views: Intensity, LBP, and Gabor. $\textbf{HandWritten}$\footnote{https://archive.ics.uci.edu/ml/datasets/Multiple+Features} includes 2,000 digit samples (200 per class from ‘0’ to ‘9’), with six views: Fourier coefficients (FOU), profile correlations (FAC), Karhunen–Loeve coefficients (KAR), pixel averages, Zernike moments (ZER), and morphological features (MOR). $\textbf{ALOI}$\footnote{https://elki-project.github.io/datasets/multi\_view} contains 10,158 images of 1,000 small objects, each represented by four sets of deep features. $\textbf{NUS-WIDE-OBJECT}$\footnote{https://lms.comp.nus.edu.sg/wp-content/uploads/2019/} (NUS) has 30,000 images spanning 31 categories, described using five views: Color Histogram, block-wise Color Moments, Color Correlogram, Edge Direction Histogram, and Wavelet Texture. $\textbf{MOSI}$\footnote{https://multicomp.cs.cmu.edu/resources/cmu‑mosi‑dataset/} is a sentiment analysis dataset with 1,284 samples classified as positive, negative, or neutral, using three views: text, vision, and audio. $\textbf{Food-101}$\footnote{https://visiir.isir.upmc.fr} is a large-scale multi-modal food recognition benchmark comprising 87,696 samples distributed across 101 food categories. Each sample is represented through three distinct modalities, namely: class-level textual descriptions (class-text), detailed textual instructions or ingredients (dect-text), and visual information (image). 
As for \textbf{conflictive datasets}, we construct them based on the above normal datasets by injecting controlled inconsistency into the test samples. Specifically, we simulate view-level conflicts by introducing two types of perturbations: for noise views, we add Gaussian noise with varying standard deviations to one or more views, creating noisy representations; additionally, for unaligned views, we deliberately alter the information in a randomly chosen view such that its semantic content becomes inconsistent with the true label, resulting in label-view misalignment. These modifications yield challenging test scenarios that reflect real-world uncertainty and view disagreement, providing a rigorous benchmark for evaluating the robustness and reliability of multi-view learning methods under conflictive conditions.

\subsection{Compared methods}
The compared baselines based on feature fusion are summarized as follows: (1) \textbf{EDL}~\cite{sensoy2018evidential} utilizes Dirichlet distributions to model prediction uncertainty explicitly, enabling confidence-aware classification. (2) \textbf{DCCAE}~\cite{wang2015deep} is a classical multi-view learning method that leverages autoencoders to learn a shared latent representation by maximizing canonical correlations. (3) \textbf{CPM-Nets} ~\cite{zhang2019cpm} is a state-of-the-art feature fusion framework for multi-view learning. It aims to learn a unified and flexible representation by effectively modeling the intricate correlations among multiple views. (4) \textbf{DUA-Nets}~\cite{geng2021uncertainty} incorporate uncertainty modeling via reversal networks, dynamically integrating information from multiple views into a united representation. (5) \textbf{TMC}~\cite{han2021trusted} focuses on uncertainty-aware classification by introducing mechanisms to provide reliable and trustworthy predictions. (6) \textbf{TMDL-OA}~\cite{Liu_Yue_Chen_Denoeux_2022} is a state-of-the-art decision fusion method based on evidential deep neural networks. It incorporates a consistency measure loss to enhance the reliability of multi-view learning outcomes. (7) \textbf{RCML}~\cite{xu2024reliable} addresses multi-view reliability by modeling both shared and specific view-level opinions and aggregating them using a conflict-aware opinion mechanism. (8) \textbf{RMVC}~\cite{yue2024evidential} enhances robustness under adversarial perturbations by measuring evidential dissonance and guiding fusion through dissonance-aware belief integration. (9) \textbf{TUNED} ~\cite{huang2024trusted} enhances robust fusion by integrating local and global feature-neighborhood structures, employing a shared evidence extractor and selective Markov random field to mitigate uncertainty and view conflicts.

\subsection{Further Analysis}
\begin{table}
    \centering
    \small
    \begin{tabular}{p{0.95cm}p{0.78cm}p{0.78cm}p{0.78cm}p{0.78cm}p{0.78cm}p{0.78cm}}
        \toprule
        \multirow{2}{*}{Methods} & \multicolumn{2}{c}{HandWritten} & \multicolumn{2}{c}{PIE} & \multicolumn{2}{c}{MOSI} \\
        \cmidrule(r){2-3} \cmidrule(r){4-5} \cmidrule(r){6-7}
         & Time & FLOPs & Time & FLOPs & Time & FLOPs \\
        \midrule
        RCML  & 0.0190s & 4.36M & 0.0619s & 31.5M & 0.2327s & 30.56M\\
        TUNED & 0.0769s & 394.9G & 0.6587s & 10.14G & 3.2494s & 328.06G \\
        Ours  & 0.0216s & 13.1M & 0.0689s & 47.26M & 0.4855s & 48.9M \\
        \bottomrule
    \end{tabular}
    \vspace{-1em}
    \caption{Average training time per epoch and FLOPs on three prevalent multi-view datasets.}
    \label{tab:time}
    \vspace{-1em}
\end{table}


\paragraph{Computational Time.}

We compare the average per-epoch training time and FLOPs of RCML, TUNED, and our method on three benchmark datasets (\textbf{Table~\ref{tab:time}}). Our method achieves training efficiency comparable to RCML and is significantly faster than TUNED, which aligns well with our theoretical complexity analysis. The improved efficiency primarily arises because our approach eliminates the need for expensive pairwise similarity computations and graph-based operations commonly used in other methods. Instead, it employs a lightweight prototype-based framework that efficiently captures class-level representations, greatly reducing the computational overhead. This design not only accelerates the training process but also scales better with increasing data size and number of views, making our method more practical for large-scale multi-view learning scenarios. Consequently, the combination of competitive accuracy and reduced training time and FLOPs demonstrates the effectiveness and scalability of our proposed approach.

\paragraph{Parameters analysis}

\textbf{Figure~\ref{fig10}} present the sensitivity analysis of key parameters on the PIE, Handwritten, and MOSI datasets under both normal and conflict settings.

In \textbf{Figure~\ref{fig10}}, we analyze the effects of the size of the selected KNN feature set $R$ and hyperparameters $\alpha$, $\beta$ and $\gamma$, which linearly weight the different components of our loss function. Each of these loss-related parameters plays a critical role in balancing the contributions of respective loss terms, thereby directly influencing the final performance. This confirms the necessity of their inclusion and appropriate tuning within the overall objective. The parameter $|R|$ controls the neighborhood scope during the loss calculation, reflecting the number of nearest neighbors sampled from the feature space. A larger $|R|$ value causes the model to consider a more global neighborhood structure, potentially smoothing over local variations, whereas a smaller $|R|$ value places more emphasis on local, fine-grained neighborhood relationships. The results indicate that moderate values of $|R|$ achieve a desirable balance, capturing sufficient neighborhood information without diluting locality. For the regularization parameter $\xi$, we experimented with values from the set $\{1, 10^1, 10^2\}$. Due to the substantial differences among datasets, the choice of $\xi$ is largely empirical, and we do not provide further details here.

In conclusion, the model’s robustness to a range of prototype sample sizes reflects the flexibility of the embedded neighborhood mechanism, but fine-tuning this parameter can further enhance performance by striking the right trade-off between local and global representation.

\end{document}